\documentclass{article}
\usepackage[utf8]{inputenc}
\usepackage{xcolor}
\usepackage{amsmath}
\usepackage{graphicx}

\makeatletter
\usepackage{ijcai18}

\usepackage{times}
\usepackage{soul}
\usepackage{url}
\usepackage[small]{caption}
\usepackage{amsfonts}
\usepackage{array}
\usepackage{multirow}
\usepackage{longtable}
\usepackage{subfigure}
\usepackage{algorithm}
\usepackage{algpseudocode}

\title{Soft Filter Pruning for Accelerating Deep Convolutional Neural Networks}

\author{
Yang He$^{1,2}$, 
Guoliang Kang$^2$, 
Xuanyi Dong$^2$,
Yanwei Fu$^{3}$\thanks{Corrsponding Author}, 
Yi Yang$^{1,2*}$
\\ 
$^1$SUSTech-UTS Joint Centre of CIS, Southern University of Science and Technology\\ $^2$CAI, University of Technology Sydney\\
$^3$The School of Data Science, Fudan University\\
\{yang.he-1, guoliang.kang, xuanyi.dong\}@student.uts.edu.au, \\
yanweifu@fudan.edu.cn,
yi.yang@uts.edu.au
}
\makeatother

\begin{document}
\maketitle %%%%%%%%%%%%%%%%%%%%%%%%%%%%%%%%%%%%%%%%%%%%%%%%%%%%%%%%%%%%%%%%%%%%%%%%%%%%%%%%%%%%%%%%%%%%%%%%%%%
\begin{abstract}
%This paper proposed a Soft Filter Pruning (SFP) method to prune the filters of deep Convolutional Neural Networks (CNNs) which can thus be accelerated in the inference.
%\footnote{Code:~https://github.com/he-y/soft-filter-pruning}
This paper proposed a Soft Filter Pruning (SFP) method to accelerate the inference procedure of deep Convolutional Neural Networks (CNNs). Specifically, the proposed SFP enables the pruned filters to be updated when training the model after pruning. SFP has two advantages over previous works: (1) \textbf{Larger model capacity.} Updating previously pruned filters provides our approach with larger optimization space than fixing the filters to zero. Therefore, the network trained by our method has a larger model capacity to learn from the training data. (2) \textbf{Less dependence on the pre-trained model.} Large capacity enables SFP to train from scratch and prune the model simultaneously. In contrast, previous filter pruning methods should be conducted on the basis of the pre-trained model to guarantee their performance. Empirically, SFP from scratch outperforms the previous filter pruning methods. Moreover, our approach has been demonstrated effective for many advanced CNN architectures. Notably, on ILSCRC-2012, SFP reduces more than 42\% FLOPs on ResNet-101 with even 0.2\% top-5 accuracy improvement, which has advanced the state-of-the-art. Code is publicly available on GitHub: https://github.com/he-y/soft-filter-pruning
\end{abstract}
%%%%%%%%%%%%%%%%%%%%%%%%%%%%%%%%%%%%%%%%%%%%%%%%%%%%%%%%%%%%%%%%%%%%%%%%%%%%%%%%%%%%%%%%%%%%%%%%%%%

\section{Introduction}

The superior performance of deep CNNs usually comes from the deeper and wider architectures, which cause the prohibitively expensive computation cost.
Even if we use more efficient architectures, such as residual connection~\cite{he2016deep} or inception module~\cite{szegedy2015going}, it is still difficult in deploying the state-of-the-art CNN models on mobile devices.
For example, ResNet-152 has 60.2 million parameters with 231MB storage spaces; besides, it also needs more than 380MB memory footprint and six seconds (11.3 billion float point operations, FLOPs) to process a single image on CPU.
The storage, memory, and computation of this cumbersome model significantly exceed the computing limitation of current mobile devices.
Therefore, it is essential to maintain the small size of the deep CNN models which has relatively low computational cost but high accuracy in real-world applications.

%\textcolor{red}{{[}YW: two problems: (1) can you give a more exact estimation of normal CPU/GPU's computation power in order to show the inefficiency in real-world applications?
%(2) the size of deep model and computing speed of deep model are two things.
%small size not necessarily lower computational speed. We need to elaborate this point. {]}. }
%\iffalse
\begin{figure}[!t]
\begin{centering}
\includegraphics[width=0.48\textwidth]{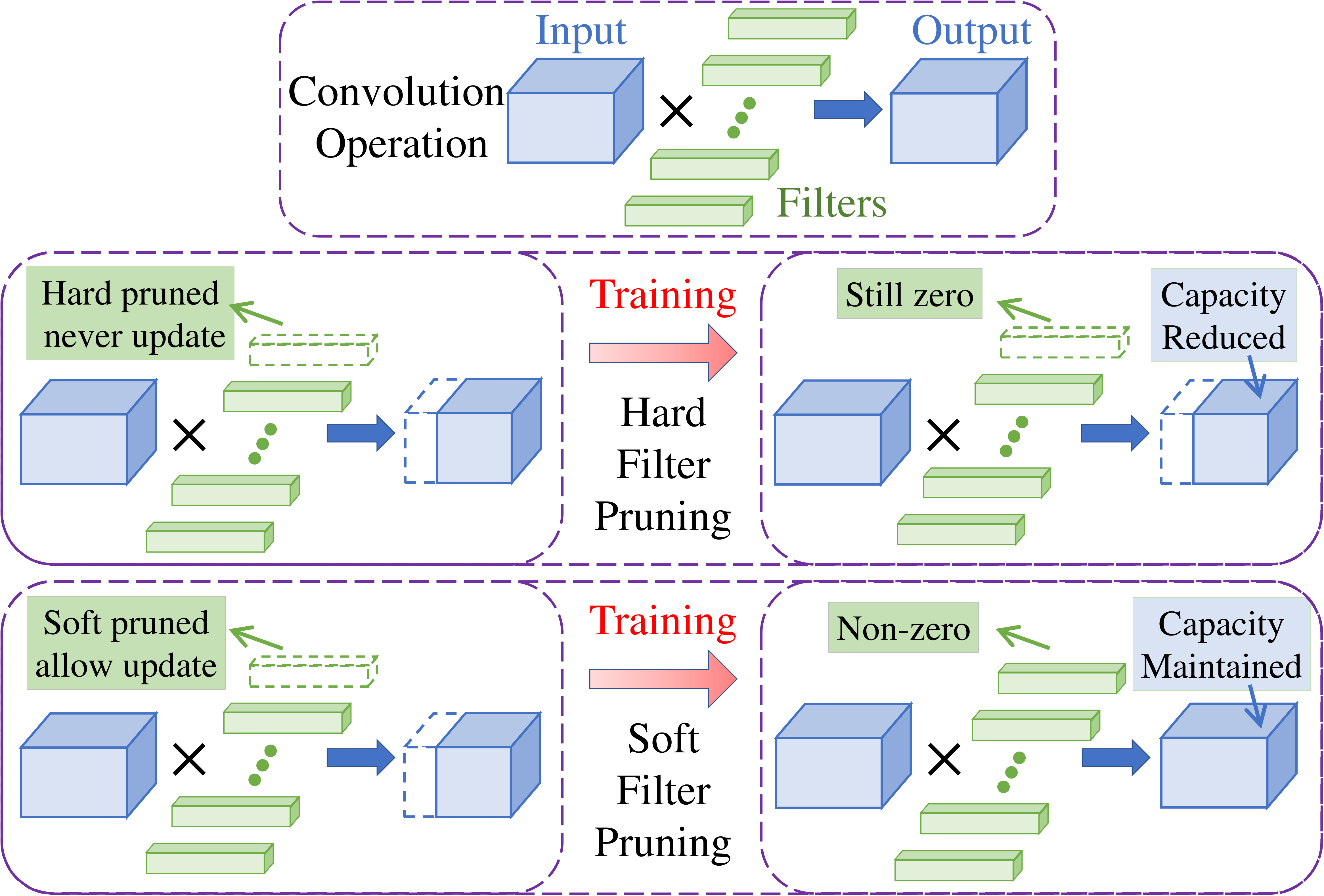} 
\par\end{centering}
\caption{
\textbf{Hard Filter Pruning}~\emph{v.s.}~\textbf{Soft Filter Pruning}.
We mark the pruned filter as the green dashed box.
For the hard filter pruning, the pruned filters are always \textbf{\emph{fixed}} during the whole training procedure.
Therefore, the model capacity is reduced and thus harms the performance because the dashed blue box is useless during training.
On the contrary, our SFP \textbf{\emph{allows}} the pruned filters to be updated during the training procedure.
In this way, the model capacity is recovered from the pruned model, and thus leads a better accuracy.
}
\label{fig:hard_soft} 
\end{figure}
%\fi
% Recent efforts are made on pruning the full deep CNN models~\cite{reed1993pruning} to reduce
% the model size and save computational cost while keeping a reasonable
% good performance. These previous works roughly
% focus on either removing the deep CNN models, 

%It is thus necessary to prune deep CNNs~\cite{reed1993pruning}.
Recent efforts have been made either on directly deleting weight values of
filters~\cite{han2015learning} (i.e., weight pruning) or totally
discarding some filters (i.e., filter pruning)~\cite{li2016pruning,He_2017_ICCV,Luo_2017_ICCV}.
However, the weight pruning may result in the unstructured sparsity
of filters, which may still be less efficient in saving the memory
usage and computational cost, since the unstructured model cannot
leverage the existing high-efficiency BLAS libraries. In contrast,
the filter pruning enables the model with structured sparsity and
more efficient memory usage than weight pruning, and thus takes full advantage of BLAS libraries to achieve a more realistic acceleration.
Therefore, the filter pruning is more advocated in accelerating the
networks.  

%%%%%%%%%%%old version
% The standard practices address this problem by either pruning the weights \cite{han2015learning} or the filters \cite{li2016pruning,He_2017_ICCV,Luo_2017_ICCV} of the deep CNN networks.
% The key part of the solution is to prune some components of trained models \cite{reed1993pruning}.
% After pruning, the fine-tuning step is still necessary to maintain a comparable performance with respect to the original full trained models.
% In general, the weight pruning usually results in the unstructured sparsity for deep CNN models.
% This approach makes the model difficult to save the memory usage and the computation cost,
% because the unstructured model cannot leverage the existing high-efficiency BLAS libraries.
% In contrast, the filter pruning enables the model with structured sparsity.
% Therefore, it can effectively leverage these libraries to achieve the realistic acceleration.
% Moreover, the memory usage would be reduced dramatically compared to weight pruning~\cite{Luo_2017_ICCV}.
% Therefore, we advocate the filter pruning in accelerating the networks.
% Furthermore, in this paper, we aim to facilitate improvements in both accuracy and acceleration for the filter pruning approach.
%%%%%%%%%%%old version

Nevertheless, most of the previous works on filter pruning
still suffer from the problems of (1) \emph{the model capacity reduction}
and (2) \emph{the dependence on pre-trained model}.
Specifically, as shown in Fig.~\ref{fig:hard_soft}, most previous works conduct the
``hard filter pruning'', which directly delete the pruned filters. The discarded filters will reduce the model capacity
of original models, and thus inevitably harm the performance. Moreover,
to maintain a reasonable performance with respect to the full models,
previous works~\cite{li2016pruning,He_2017_ICCV,Luo_2017_ICCV} always
fine-tuned the hard pruned model after pruning the filters of a pre-trained model,
which however has low training efficiency and often requires much
more training time than the traditional training schema. 

%%%%%%%%%%%old version
% Despite the above advantages of the filter pruning,
% there exist two issues for most of the filter pruning algorithms:
% 1) \emph{the model capacity reduction} and 2) \emph{the low training efficiency}.
% The previous methods always fix the pruned filters and never update them~\cite{li2016pruning,He_2017_ICCV,Luo_2017_ICCV}, see hard filter pruning in Figure~\ref{fig:hard_soft}.
% In the convolutional layer, one filter corresponds to one feature map.
% Therefore, for the hard filter pruning, the feature maps of the pruned filters are useless during the following training procedure.
% In this way, the model capacity is reduced compared to the original architecture, and thus inevitably harm the performance.
% Moreover, most previous methods require to firstly train the original model, and then iterate over two steps, \emph{i.e.}, greedily pruning filters and fine-tuning the model~\cite{li2016pruning,He_2017_ICCV,Luo_2017_ICCV}.
% This manner costs much more training time than the traditional training schema, especially for deep CNN models.
%%%%%%%%%%%old version

To address the above mentioned two problems, we propose a novel Soft
Filter Pruning (SFP) approach. The SFP dynamically prunes the filters
in a soft manner. Particularly, before first training epoch, the filters of almost all layers with small $\ell_{2}$-norm are selected
and set to zero. Then the training data is used to update the pruned
model. Before the next training epoch, our SFP will prune
a new set of filters of small $\ell_{2}$-norm.
These training process is continued until converged. Finally, some
filters will be selected and pruned without further updating. The
SFP algorithm enables the compressed network to have a larger model capacity,
and thus achieve a higher accuracy than others.

%%%%%%%%%%%old version
% To address the above mentioned two problems, we propose a novel soft filter pruning approach.
% It allows the pruned filters to recover during the training procedure.
% Specifically, after each training epoch, we first select unimportant filters by a $\ell_{2}$-norm selection criteria. We set these selected filters to zero, then we train the model for one more epoch before we prune some filters again. To be noticed, these "zeroized" filters can still update during training.
% This provides our approach with larger optimization space then fixing the filters to zero.
% Therefore, we enable the network with a larger model capacity, and thus can achieve a higher accuracy than others.
%%%%%%%%%%%old version

\textbf{Contributions.} We highlight three contributions:
(1) We propose SFP to allow the pruned filters to be updated during the training procedure.
This soft manner can dramatically maintain the model capacity and thus achieves the superior performance.
(2)~Our acceleration approach can train a model from scratch and achieve better performance compared to the state-of-the-art.
In this way, the fine-tuning procedure and the overall training time is saved.
Moreover, using the pre-trained model can further enhance the performance of our approach to advance the state-of-the-art in model acceleration.
(3)~The extensive experiment on two benchmark datasets demonstrates the effectiveness and efficiency of our SFP.
We accelerate ResNet-110 by two times with about 4\% relative accuracy improvement on CIFAR-10, and also achieve state-of-the-art results on ILSVRC-2012.

\section{Related Works}

Most previous works on accelerating CNNs can be roughly divided into three
categories, namely, \emph{matrix decomposition}, \emph{low-precision
weights}, and \emph{pruning}.
In particular, the \emph{matrix decomposition} of deep CNN tensors is approximated by the product of two low-rank
matrices~\cite{jaderberg2014speeding,zhang2016accelerating,tai2015convolutional}.
This can save the computational cost.
Some works \cite{zhu2016trained,zhou2017incremental} focus on compressing the CNNs by using \emph{low-precision weights}.
\emph{Pruning}-based approaches aim to remove the unnecessary connections of the neural network \cite{han2015learning,li2016pruning}. Essentially,
the work of this paper is based on the idea of pruning techniques;
and the approaches of matrix decomposition and low-precision weights
are orthogonal but potentially useful here – it may be still worth
simplifying the weight matrix after pruning filters, which would
be taken as future work. 
% In this way, they can reduce the computation cost as well as the memory usage.

\textbf{Weight Pruning.}
Many recent works~\cite{han2015learning,han2015deep,guo2016dynamic} pruning weights of neural network resulting in small models.
For example,
\cite{han2015learning} proposed an iterative weight pruning method by discarding the small weights whose values are below the threshold.
% The pruned weights are updated by a fine-tuning process which however requires high number of iteration steps to be converged, and thus is relatively less practical in real-world application.
\cite{guo2016dynamic} proposed the dynamic network surgery to reduce the training iteration while maintaining a good prediction accuracy.
%\cite{guo2016dynamic} proposed
%dynamic network surgery which is a network compression method.
\cite{wen2016learning,lebedev2016fast} leveraged the sparsity property of feature maps or weight parameters to accelerate the CNN models.
A special case of weight pruning is neuron pruning.
% \cite{hu2016network} evaluated the importance of neurons by measuring the sparsity of ReLU activations.
% percentage of zero activations of a neuron after the ReLU mapping.
% To this end, \cite{wen2016learning} proposed the Structured Sparsity
% Learning (SSL) method to regularize filter, channel, filter shape
% and depth structures. \cite{lebedev2016fast} applied the group-sparsity
% regularization on the loss function to shrink some entire groups of
% weights towards zeros.
% Despite their success in \cite{wen2016learning,lebedev2016fast},
% it is still computational expensive in computing gradients of the additional regularization term with respect to all the weights.
% Neuron Pruning has also been considered as 
% The neurons with lower importance can be pruned.
However, pruning weights always leads to unstructured models, so the model cannot leverage the existing efficient BLAS libraries in practice. Therefore, it is difficult for weight pruning to achieve realistic speedup.
% and thus inefficient in reducing the inference time of the network.

%\vspace{0.05in}

\textbf{Filter Pruning.} Concurrently with our work, some
filter pruning strategies \cite{li2016pruning,Liu_2017_ICCV,He_2017_ICCV,Luo_2017_ICCV}
have been explored. Pruning the filters leads to the removal of the
corresponding feature maps. This not only reduces the storage usage
on devices but also decreases the memory footprint consumption to
accelerate the inference. \cite{li2016pruning} uses $\ell_{1}$-norm
to select unimportant filters and explores the sensitivity of layers
for filter pruning. \cite{Liu_2017_ICCV} introduces $\ell_{1}$
regularization on the scaling factors in batch normalization (BN)
layers as a penalty term, and prune channel with small scaling factors
in BN layers. \cite{molchanov2016pruning} proposes a Taylor expansion
based pruning criterion to approximate the change in the cost function
induced by pruning. \cite{Luo_2017_ICCV} adopts the statistics information
from next layer to guide the importance evaluation of
filters. \cite{He_2017_ICCV} proposes a LASSO-based channel selection strategy, and a least square reconstruction algorithm to prune filers.
However, for all these filter pruning methods, the representative
capacity of neural network after pruning is seriously
affected by smaller optimization space.

\textbf{Discussion.}
To the best of our knowledge, there is only one approach that uses the soft manner to prune weights~\cite{guo2016dynamic}.
We would like to highlight our advantages compared to this approach as below:
(1) Our SPF focuses on the filter pruning, but they focus on the weight pruning.
As discussed above, weight pruning approaches lack the practical implementations to achieve the realistic acceleration.
(2) \cite{guo2016dynamic} paid more attention to the model compression, whereas our approach can achieve both compression and acceleration of the model.
(3) Extensive experiments have been conducted to validate the effectiveness of our proposed approach both on large-scale datasets and the state-of-the-art CNN models.
In contrast, \cite{guo2016dynamic} only had the experiments on Alexnet which is more redundant the advanced models, such as ResNet. 

% We have demonstrated our approach on large-scale datasets with various of CNN models.
% However, \cite{guo2016dynamic} only experiment on AlexNet, which is more redundant than the advanced models, such as ResNet. It is unclear their effect on the state-of-the-art CNN models. 

\section{Methodology}
%This section firstly gives the problem setup in Sec. \ref{Preliminary};then the details of SFP algorithm is fully developed in Sec. \ref{Soft Filter Pruning}. Finally, some theoretical analysis is given in Sec. \ref{Calculation Reduction After Pruning}.
% In this section, we will give a comprehensive introduction to our SFP and
% present its implementation details.

\subsection{Preliminaries}\label{Preliminary}

We will formally introduce the symbol and annotations in this section.
The deep CNN network can be parameterized by $\{\mathbf{W}^{(i)}\in\mathbb{R}^{N_{i+1}\times N_{i}\times K\times K},1\leq i\leq L\}$
%\footnote{Fully-connected layers equal to convolutional layers with $k=1$}.
$\mathbf{W}^{(i)}$ denotes a matrix of connection weights in the $i$-th layer.
$N_{i}$ denotes the number of input channels for the $i$-th convolution layer.
$L$ denotes the number of layers.
The shapes of input tensor $\mathbf{U}$ and output tensor $\mathbf{V}$ are $N_{i} \times H_{i}\times W_{i} $ and $N_{i+1} \times H_{i+1}\times W_{i+1} $, respectively.
The convolutional operation of the $i$-th layer can be written as:

\vspace{-3mm}
{\small
\begin{align}
\label{eq:1}
\mathbf{V}_{i,j}=\mathcal{F}_{i,j}\ast\mathbf{U} ~&~ \mathbf{for}~1 \leq j \leq N_{i+1},
\end{align}
}
\vspace{-3mm}

\noindent where $\mathcal{F}_{i,j} \in \mathbb{R}^{N_{i}\times K\times K}$ represents the $j$-th filter of the $i$-th layer.
$\mathbf{W}^{(i)}$ consists of $\{\mathcal{F}_{i,j}, 1 \leq j \leq N_{i+1}\}$.
% and $\mathcal{F}_{i,j}^{k}$ represents the $k$-th kernel of $\mathcal{F}_{i,j}$.
The $\mathbf{V}_{i,j}$ represents the $j$-th output feature map of the $i$-th layer.
% which is composed by  kernels
% $\mathcal{K}$. Applying $N_{i+1}$ filters
% $\mathcal{F}_{i,j}\in\mathbb{R}^{N_{i}\times k\times k}$ on the $N_{i}$
% input channels generates $N_{i+1}$ feature maps.

Pruning filters can remove the output feature maps.
In this way, the computational cost of the neural network will reduce remarkably.
Let us assume the pruning rate of SFP is $P_{i}$ for the $i$-th layer.
The number of filters of this layer will be reduced from $N_{i+1}$ to $N_{i+1}(1-P_{i})$,
thereby the size of the output tensor $\mathbf{V}_{i,j}$ can be reduced to ${N_{i+1}(1-P_{i})\times H_{i+1} \times W_{i+1}}$.
As the output tensor of $i$-th layer is the input tensor of $i+1$-th layer,
we can reduce the input size of $i$-th layer to achieve a higher acceleration ratio.

\subsection{Soft Filter Pruning (SFP)}

\label{Soft Filter Pruning} 
\begin{figure*}[!t]
\begin{centering}
\includegraphics[width=1\textwidth]{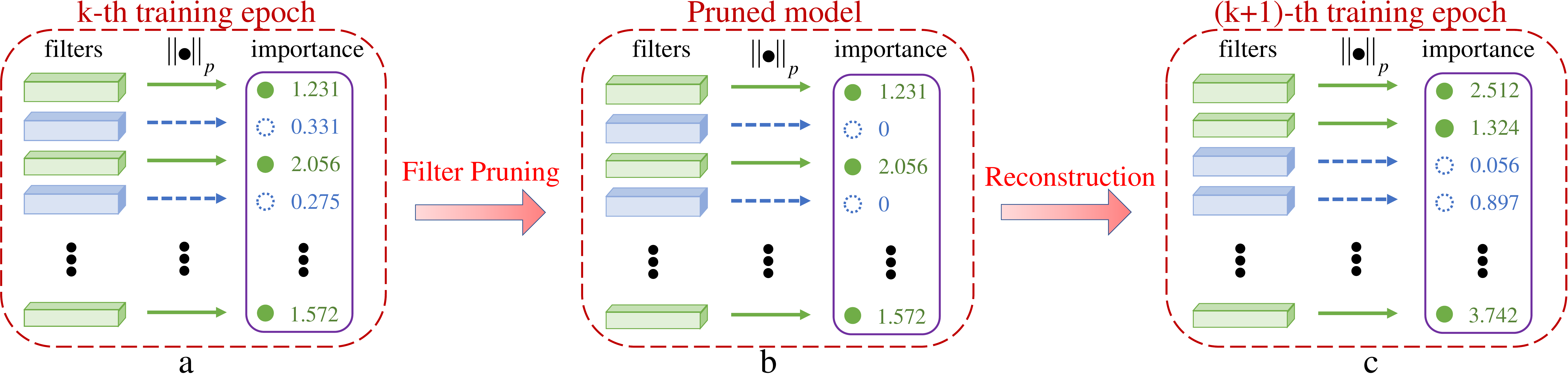} 
\par\end{centering}
\caption{
Overview of SFP.
At the end of each training epoch, we prune the filters based on their importance evaluations.
The filters are ranked by their $\ell_{p}$-norms (purple rectangles) and the small ones (blue circles) are selected to be pruned. 
After filter pruning, the model undergoes a reconstruction process 
where pruned filters are capable of being reconstructed (i.e., updated from zeros)
by the forward-backward process. \textbf{(a)}:~filter instantiations before pruning. \textbf{(b)}:~filter instantiations after pruning. \textbf{(c)}:~filter instantiations after reconstruction.
}
\label{fig:soft_pruning} 
\end{figure*}

Most of previous filter pruning works \cite{li2016pruning,Liu_2017_ICCV,He_2017_ICCV,Luo_2017_ICCV}
compressed the deep CNNs in a hard manner. We call them as the hard
filter pruning.
Typically, these algorithms
firstly prune filters of a single layer of a pre-trained model and fine-tune the pruned model to complement the degrade of the performance. Then they prune the next layer and fine-tune the model again until the last layer of the model is pruned.
However, once filters are pruned, these approaches will not update these filters again.
Therefore, the model capacity is drastically reduced due to the removed filters; and such a hard pruning manner affects the performance of the compressed models negatively. 

As summarized in Alg.~\ref{alg:SFP}, the proposed SFP algorithm can dynamically remove the filters in a soft manner. Specifically, the key is to keep updating the pruned filters in the training stage.
Such an updating manner brings several benefits. It not only keeps the model capacity of the compressed deep CNN models as the original models, but also avoids the greedy layer by layer pruning procedure and enable pruning almost \emph{all~layers} at the same time.
More specifically, our
approach can prune a model either in the process of training from
scratch, or a pre-trained model. In each training epoch, the full
model is optimized and trained on the training data. After each epoch, the
$\ell_{2}$-norm of all filters are computed for each weighted layer
and used as the criterion of our filter selection strategy.
Then we will prune the selected filters by setting the corresponding filter weights as zero, which is followed by next training epoch.
Finally, the original deep CNNs are pruned into a compact and efficient model.
The details of SFP is illustratively explained in Alg.~\ref{alg:SFP}, which can be divided into the following four steps.

\begin{algorithm}[t]
\caption{Algorithm Description of SFP}
\label{alg:SFP}

\begin{algorithmic}
\Require training data: $\mathbf{X}$, pruning rate: $P_{i}$ \\the model with parameters $\mathbf{W} = \{\mathbf{W} ^{(i)}, 0\leq i \leq L\}$.       
\State Initialize the model parameter $\mathbf{W}$   
\For{$epoch=1$; $epoch \leq epoch_{max}$; $epoch++$}
	\State Update the model parameter $\mathbf{W}$ based on $\mathbf{X}$
	\For{$i=1$; $i \leq L $; $i++$}          
		\State Calculate the $\ell_2$-norm for each filter $\|\mathcal{F}_{i,j}\|_2, 1 \leq j \leq N_{i+1}$        
		\State Zeroize $N_{i+1}P_i$ filters by $\ell_2$-norm filter selection
	\EndFor
\EndFor
\State Obtain the compact model with parameters $\mathbf{W} ^{*}$ from $\mathbf{W}$
\Ensure The compact model and its parameters $\mathbf{W} ^{*}$
\end{algorithmic} 
\end{algorithm}

%%%%%%%%%%%%%%%%%%%%%%%%%%%%%%%%%%%%%%%%%%%%%%%%%%%%%%%%%%%%%%%%%%%%%%%%%%%%%%%%%%%%%%%%%%%%%%%%%%%

\textbf{Filter selection.}
We use the $\ell_{p}$-norm to evaluate the importance of each filter as Eq.~\eqref{eq:p-norm}.
%%%%%%%%%%%%%%%%%%%%%%old version
%The filters with the lower $\ell_{p}$-norm have less influence to the final CNN prediction. 
% Therefore, such filters are less important than others with the higher $\ell_{p}$-norm.
% In our approach,
In general, the convolutional results of the filter with the smaller
$\ell_{p}$-norm lead to relatively lower activation values; and
thus have a less numerical impact on the final prediction of deep CNN
models. In term of this understanding, such filters of small $\ell_{p}$-norm
will be given high priority of being pruned than those of higher $\ell_{p}$-norm.
Particularly, we use a pruning rate $P_{i}$ to select $N_{i+1}P_{i}$ unimportant filters for the $i$-th weighted layer.
In other words, the lowest $N_{i+1}P_{i}$ filters are selected, e.g., the blue filters in Fig.~\ref{fig:soft_pruning}.
In practice, $\ell_{2}$-norm is used based on the empirical analysis.

% The $\ell_{p}$-norm is a magnitude based criteria.

\vspace{-3mm}
{\small
\begin{align}
\label{eq:p-norm}
\|\mathcal{F}_{i,j}\|_{p}=\sqrt[p]{\sum_{n=1}^{N_{i}}\sum_{k_{1}=1}^{K}\sum_{k_{2}=1}^{K}\left|\mathcal{F}_{i,j} (n,k_{1},k_{2})\right|^{p}},
\end{align}
}
\vspace{-2mm}

% When $p=2$, we get the $\ell_{2}$-norm of filter. We use
% $\ell_{2}$-norm as our selection criteria of filter. Some other magnitude
% based criteria have also been proposed to evaluate the importance
% of each filter in previous literature such as $\ell_{1}$-norm of
% the filters~\cite{li2016pruning}. After comparing these two criteria,
% we find $\ell_{2}$-norm works better than $\ell_{1}$-norm, so we
% choose the $\ell_{2}$-norm as our selection criteria. The detailed
% experiment is shown in section~\ref{1-norm and 2-norm}.
% 
% We use the $\ell_{2}$-norm of filters to measure the importance of filters.
% Specifically,  and the detailed explanation is given in section~\ref{Selection of Filter Within A layer}.
% For $N_{i+1}$ filters within \textit{i}th layer, and the pruning rate is $P_{i}$,
% then $N_{i+1}\times P_{i}$ filters which has the lowest $\|\mathcal{F}_{i,j}\|_{2}$ value is to be pruned in this layer.
% As shown in Fig.~\ref{fig:soft_pruning},
% the blue filters indicate the filters that have the smallest $\|\mathcal{F}_{i,j}\|_{2}$,
% and these filters should be pruned.

\textbf{Filter Pruning.}
We set the value of selected $N_{i+1}P_{i}$ filters to zero (see the filter pruning step in Fig.~\ref{fig:soft_pruning}).
This can temporarily eliminate their contribution to the network output.
Nevertheless, in the following training stage, we still allow these selected filters to be updated,
in order to keep the representative capacity and the high performance
of the model. 
%old version
% update.
% In this way, we can maintain the representative capacity of the neural network.
% Therefore, the model can potentially obtain the higher performance due to the maintained capacity.

% \DXYB{Hard filter pruning always use a layer by layer
% or even filter by filter greedy pruning scheme, which means a huge
% number of pruning and training cycles are needed. On the contrary,
% our SFP prunes all the selected filters in all layers at the same
% time, thus saving lots of training time especially for deep models.}
% After SFP, the $(N_{i+1}\times P_{i})\times H_{i+1}\times W_{i+1}$
% feature maps corresponding to the pruned filters are all change to
% zero, thus we have a compact model to accelerate the neural network
% inference.

In the filter pruning step, we simply prune \emph{all} the weighted layers at the same time.
In this way, we can prune each filter in parallel, which would cost negligible computation time.
In contrast, the previous filter pruning methods always conduct layer by layer greedy pruning.
After pruning filters of one single layer, existing methods always require training to converge the network~\cite{Luo_2017_ICCV,He_2017_ICCV}.
This procedure cost much extra computation time, especially when the depth increases.
Moreover, we use the \emph{same} pruning rate for \emph{all} weighted layers.
Therefore, we need only one hyper-parameter $P_{i}=P$ to balance the acceleration and accuracy.
This can avoid the inconvenient hyper-parameter search or the complicated sensitivity analysis~\cite{li2016pruning}. As we allow the pruned filters to be updated, the model has a large model capacity and becomes more flexible and thus can well balance the contribution of each filter to the final prediction.

%The large model capacity allow the model to learn from training data quickly, thus the influence of all layer pruning will be eliminate soon.

%Why can SPF prune all layers by the same pruning rate at the same time?
%Intuitively, this is due to the large model capacity of the model trained by SPF.
%As we allow the pruned filters to be updated, the model has large model capacity and becomes more flexible and thus can well balance the contribution of each filter to the final prediction.
%Therefore, we may have a better performance than previous greedy method. Besides, less pruning and \DXYB{reconstruction} cycles are needed if we prune all layers together.
%Furthermore, we just have one hyper-parameter which is directly related to the acceleration, that is, the pruning rate $P_{i}=P$.

%\textbf{(3) Rehabilitation Training.}%[other options: Rehabilitation, Training-by-recovery]}
{\bf Reconstruction.}
After the pruning step, we train the network for one epoch to reconstruct the pruned filters.
As shown in Fig.~\ref{fig:soft_pruning}, the pruned filters are updated to non-zero by back-propagation.
In this way, SFP allows the pruned model to have the same capacity as the original model during training.
%Therefore, we can potentially obtain the similar performance compared to the original model.
In contrast, hard filter pruning decreases the number of feature maps.
The reduction of feature maps would dramatically reduce the model capacity, and further harm the performance.
Previous pruning methods usually require a pre-trained model and then fine-tune it.
However, as we integrate the pruning step into the normal training schema, our approach can train the model from scratch.
Therefore, the fine-tuning stage is no longer necessary for SFP.
As we will show in experiments, the network trained from scratch by SFP can obtain the competitive results with the one trained from a well-trained model by others.
By leveraging the pre-trained model, SFP obtains a much higher performance and advances the state-of-the-art.
%thus for a scratch model which has not learned the feature of training
%set yet, the reduction of feature map dimension is particularly fatal to the performance.

\textbf{Obtaining Compact Model.}
SFP iterates over the filter selection, filter pruning and reconstruction steps.
After the model gets converged, we can obtain a sparse model containing many ``zero filters".
One ``zero filter" corresponds to one feature map.
The features maps, corresponding to those ``zero filters", will always be zero during the inference procedure.
There will be no influence to remove these filters as well as the corresponding feature maps.
Specifically, for the pruning rate $P_{i}$ in the $i$-th layer, only $N_{i+1}(1-P_{i})$ filters are non-zero and have an effect on the final prediction.
Consider pruning the previous layer, the input channel of $i$-th layer is changed from $N_{i}$ to $N_{i}(1-P_{i-1})$.
%$\mathcal{F}_{i,j}$ have the contribution to the final result of the neural network.
We can thus re-build the $i$-th layer into a smaller one.
%new weighted layer which has a dimension of ${N_{i+1}(1-P_{i})\times N_{i}\times K\times K}$.
Finally, a compact model $\{\mathbf{W^{*}}^{(i)}\in\mathbb{R}^{N_{i+1}(1-P_{i})\times N_{i}(1-P_{i-1})\times K\times K}\}$ is obtained.

\begin{table*}[th]
\setlength{\tabcolsep}{1.2mm}
\small  	
\centering  	
\begin{tabular}{|c|c c c c c c c|}  		
\hline 		
Depth & Method &Fine-tune? &Baseline Accu. (\%)  &Accelerated Accu. (\%)  &Accu. Drop (\%) & FLOPs & Pruned FLOPs(\%)       
\\ \hline 		 	        
\multirow{4}{*}{20}         &~\cite{Dong_2017_CVPR} &N       & 91.53        & 91.43 &0.10 &	3.20E7	  & 20.3	 \\          
& Ours(10\%) & N       & \textbf{92.20 $\pm$  0.18}       &  \textbf{92.24 $\pm$ 0.33}  &\textbf{-0.04} & 3.44E7 	&15.2\\         
& Ours(20\%) & N       & \textbf{92.20 $\pm$  0.18}       & 91.20 $\pm$  0.30  & 1.00	& 2.87E7 	&29.3  \\         
& Ours(30\%) & N       & \textbf{92.20 $\pm$  0.18}       &  90.83 $\pm$  0.31 & 1.37&  \textbf{2.43E7} &\textbf{42.2}  \\\hline           

\multirow{4}{*}{32} &~\cite{Dong_2017_CVPR} &N       &92.33         & 90.74 &1.59 &	4.70E7	  & 31.2	 \\          
& Ours(10\%)   & N    & \textbf{92.63 $\pm$  0.70}       &   \textbf{93.22 $\pm$ 0.09} & \textbf{-0.59}&  5.86E7 &14.9  \\         
& Ours(20\%)   & N    & \textbf{92.63 $\pm$  0.70}       &  90.63 $\pm$  0.37 & 0.00 &  4.90E7 &28.8  \\   
& Ours(30\%)   & N    & \textbf{92.63 $\pm$  0.70}       &  90.08 $\pm$  0.08 & 0.55 &  \textbf{4.03E7} &\textbf{41.5} \\\hline  

\multirow{10}{*}{56} 		 &~\cite{li2016pruning} &N     & 93.04                   & 91.31  &1.75 &9.09E7  &27.6		 \\         
&~\cite{li2016pruning} &Y     & 93.04          & 93.06 &-0.02 &9.09E7 & 27.6 \\	
&~\cite{He_2017_ICCV} &N       & 92.80                      & 90.90 &1.90 &	-	  & 50.0	 \\         
&~\cite{He_2017_ICCV} &Y       & 92.80                   & 91.80 &1.00 &	-	 &	50.0 	 \\         
& Ours(10\%)   &N 	&\textbf{93.59 $\pm$ 0.58}   & \textbf{93.89 $\pm$ 0.19} & \textbf{-0.30}&  1.070E8 &14.7 \\        
& Ours(20\%)   &N 	&\textbf{93.59 $\pm$ 0.58}   & 93.47 $\pm$ 0.24 & 0.12 &8.98E7 &28.4\\       
& Ours(30\%)   &N 	&\textbf{93.59 $\pm$ 0.58}  & 93.10 $\pm$ 0.20 & 0.49  &  {7.40E7} &41.1 \\        
& Ours(30\%)   &Y 	&\textbf{93.59 $\pm$ 0.58}   & 93.78 $\pm$ 0.22 & -0.19 & {7.40E7} &41.1  \\
& Ours(40\%)   &N 	&\textbf{93.59 $\pm$ 0.58}  & 92.26 $\pm$ 0.31 & 1.33   &  \textbf{5.94E7} &\textbf{52.6} \\        
& Ours(40\%)   &Y 	&\textbf{93.59 $\pm$ 0.58}   & 93.35 $\pm$ 0.31 &0.24   & \textbf{5.94E7} &\textbf{52.6} \\ \hline          

\multirow{7}{*}{110}          		&~\cite{li2016pruning} &N  &93.53     & 92.94      &  0.61              & 1.55E8 	&38.6 	 \\          
&~\cite{li2016pruning} &Y    & 93.53    & 93.30  &0.20 	&1.55E8 	&38.6\\ 
&~\cite{Dong_2017_CVPR} &N       & 93.63         & 93.44 &0.19 &	-	  & 34.2 	 \\          
& Ours(10\%)   &N  & \textbf{93.68 $\pm$ 0.32}	& 93.83 $\pm$ 0.19  & -0.15 &  2.16E8 &14.6 \\  		
& Ours(20\%)   &N & \textbf{93.68 $\pm$ 0.32}	&\textbf{93.93 $\pm$ 0.41} & \textbf{-0.25 }&  1.82E8 &28.2 \\  		
& Ours(30\%)   &N & \textbf{93.68 $\pm$ 0.32} 	& 93.38 $\pm$ 0.30 & 0.30& \textbf{1.50E8} &\textbf{40.8}   \\  		
& Ours(30\%)   &Y & \textbf{93.68 $\pm$ 0.32} 	& 93.86 $\pm$ 0.21 & -0.18& \textbf{1.50E8} &\textbf{40.8}   \\     \hline  	
\end{tabular}  	
\caption{Comparison of pruning ResNet on CIFAR-10.
In ``Fine-tune?'' column, ``Y'' and ``N'' indicate whether to use the pre-trained model as initialization or not, respectively.
The ``Accu.~Drop'' is the accuracy of the pruned model minus that of the baseline model, so negative number means the accelerated model has a higher accuracy than the baseline model.
A smaller number of "Accu.~Drop" is better.
} 	
\label{table:cifar10_accuracy} 
\end{table*}  

%%%%%%%%%%%%%%%%%%%%%%%%%%%%%%%%%%%%%%%%%%%%%%%%%%%%%%%%%%%%%%%%%%%%%%%%%%%%%%%%%%%%%%%%%%%%%%%%%%%

%%%%%%%%%%%%%%%%%%%%%%%%%%%%%%%%%%%%%%%%%%%%%%%%%%%%%%%%%%%%%%%%%%%%%%%%%%%%%%%%%%%%%%%%%%%%%%%%%%%

\subsection{Computation Complexity Analysis}\label{Calculation Reduction After Pruning}

\textbf{Theoretical speedup analysis.}
Suppose the filter pruning rate of
the $i$th layer is $P_{i}$, which means the $N_{i+1}\times P_{i}$
filters are set to zero and pruned from the layer, and the other $N_{i+1}\times(1-P_{i})$
filters remain unchanged, and suppose the size of the input and output
feature map of \textit{i}th layer is $H_{i}\times W_{i}$ and $H_{i+1}\times W_{i+1}$.
Then after filter pruning, the dimension of useful output feature
map of the $i$th layer decreases from $N_{i+1}\times H_{i+1}\times W_{i+1}$
to $N_{i+1}(1-P_{i})\times H_{i+1}\times W_{i+1}$. Note that
the output of $i$th layer is the input of $(i+1)$ th layer.
And we further prunes the $(i+1)$th layer with a filter pruning rate
$P_{i+1}$, then the calculation of $(i+1)$th layer is decrease
from $N_{i+2}\times N_{i+1}\times k^{2}\times H_{i+2}\times W_{i+2}$
to $N_{i+2}(1-P_{i+1})\times N_{i+1}(1-P_{i})\times k^{2}\times H_{i+2}\times W_{i+2}$. In other words, a proportion of $1-(1-P_{i+1})\times(1-P_{i})$ of the original
calculation is reduced, which will make the neural network inference
much faster.

\textbf{Realistic speedup analysis.}
In theoretical speedup analysis, other operations such as batch normalization (BN) and pooling are negligible comparing to convolution operations.
Therefore, we consider the FLOPs of convolution operations for computation complexity comparison, which is commonly used in previous work~\cite{li2016pruning,Luo_2017_ICCV}.
However, reduced FLOPs cannot bring the same level of realistic speedup because non-tensor layers (e.g., BN and
pooling layers) also need the inference time on GPU~\cite{Luo_2017_ICCV}.
In addition, the limitation of IO delay, buffer switch and efficiency of BLAS libraries also lead to the wide gap between theoretical and realistic speedup ratio.
We compare the theoretical and realistic speedup in Section~\ref{section:ILSVRC}.
%%%%%%%%%%%%%%%%%old version
% In theoretical speedup analysis, the convolutional layers dominate the overall computation cost of the CNN model.
% Other operations, \emph{e.g.}, batch normalization and pooling, cost negligible comparing to convolution operations.
% However, in practice, these operations cost much more computation time than theoretical.
% % However, reduced FLOPs cannot bring the same level of realistic speedup because non-tensor layers ({\it e.g.}, batch normalization and pooling layers) also need the inference time on GPU. 
% The limitation of IO delay, buffer switch and efficiency of BLAS libraries also lead to the wide gap between theoretical and realistic speedup ratio.
% We will compare the theoretical and realistic speedup in Section~\ref{section:ILSVRC}.

%%%%%%%%%%%%%%%%%%%%%%%%%%%%%%%%%%%%%%%%%%%%%%%%%%%%%%%%%%%%%%%%%%%%%%%%%%%%%%%%%%%%%%%%%%%%%%%%%%%

\section{Evaluation and Results}\label{Experiment}

%%%%%%%%%%%%%%%%%%%%%%%%%%%%%%%%%%%%%%%%%%%%%%%%%%%%%%%%%%%%%%%%%%%%%%%%%%%%%%%%%%%%%%%%%%%%%%%%%%%
% In this section, we conduct experiments on two benchmark datasets
% to validate the effectiveness of our acceleration method.
%%%%%%%%%%%%%%%%%%%%%%%%%%%%%%%%%%%%%%%%%%%%%%%%%%%%%%%%%%%%%%%%%%%%%%%%%%%%%%%%%%%%%%%%%%%%%%%%%%%

\subsection{Benchmark Datasets and Experimental Setting}

Our method is evaluated on two benchmarks: CIFAR-10~\cite{krizhevsky2009learning} and ILSVRC-2012~\cite{russakovsky2015imagenet}.
The CIFAR-10 dataset contains 50,000 training images and 10,000 testing images, which are categorized into 10 classes.
ILSVRC-2012 is a large-scale dataset containing 1.28 million training images and 50k validation images of 1,000 classes.
%As discussed in \cite{Luo_2017_ICCV,He_2017_ICCV,Dong_2017_CVPR}, ResNet is less redundancy, and thus more difficult to get accelerated than VGGNet~\cite{simonyan2014very}.
%Therefore, we focus on pruning the challenging ResNet model.
Following the common setting in~\cite{Luo_2017_ICCV,He_2017_ICCV,Dong_2017_CVPR}, we focus on pruning the challenging ResNet model in this paper.
SFP should also be effective on different computer vision tasks, such as~\cite{kang2017shakeout,ren2015faster,dong2018sbr,shen2018biblosan,yang2010image,shen2018disan,dong2017dual}, and we will explore this in future.

In the CIFAR-10 experiments, we use the default parameter setting as~\cite{he2016identity} and follow the training schedule in~\cite{zagoruyko2016wide}.
On ILSVRC-2012, we follow the same parameter settings as~\cite{he2016deep,he2016identity}.
We use the same data argumentation strategies with PyTorch official examples~\cite{paszke2017automatic}.
%\footnote{https://github.com/pytorch/examples/tree/master/imagenet}.

We conduct our SFP operation at the end of every training epoch.
For pruning a scratch model, we use the normal training schedule.
For pruning a pre-trained model, we reduce the learning rate by 10 compared to the schedule for the scratch model.
We run each experiment three times and report the ``mean $\pm$ std''.
We compare the performance with other state-of-the-art acceleration algorithms, e.g., \cite{Dong_2017_CVPR,li2016pruning,He_2017_ICCV,Luo_2017_ICCV}.
%%%%%%%%%%%%%%%%%%%%%%%%%%%%%%%%%%%%%%%%%%%%%%%%%%%%%%%%%%%%%%%%%%%%%%%%%%%%%%%%%%%%%%%%%%%%%%%%%%%

\subsection{ResNet on CIFAR-10}

\textbf{Settings}. For CIFAR-10 dataset, we test our SFP on ResNet-20, 32, 56 and 110. We use several different pruning rates, and also analyze the difference between using the pre-trained model and from scratch.

% For CIFAR-10 dataset, we test our SFP
% on ResNet-20, 32, 56 and 110 for several different pruning rates. Some results about pruning pre-trained model are also reported.
% All the convolutional layers of ResNet are pruned with the same pruning
% rate at the same time. 

\textbf{Results}. Tab.~\ref{table:cifar10_accuracy} shows the results.
Our SFP could achieve a better performance than the other state-of-the-art hard filter pruning methods.
For example, \cite{li2016pruning} use the hard pruning method to accelerate ResNet-110 by 38.6\% speedup ratio with 0.61\% accuracy drop when without fine-tuning.
When using pre-trained model and fine-tuning, the accuracy drop becomes 0.20\%.
However, we can accelerate the inference of ResNet-110 to 40.8\% speed-up with only 0.30\% accuracy drop without fine-tuning.
When using the pre-trained model, we can even outperform the original model by 0.18\% with about more than 40\% FLOPs reduced.

These results validate the effectiveness of SFP, which can produce a more compressed model with comparable performance to the original
model. 

\subsection{ResNet on ILSVRC-2012}
\label{section:ILSVRC}

%%%%%%%%%%%%%%%%%%%%%%%%%%%%%%%%%%%%%%%%%%%%%%%%%%%%%%%%%%%%%%%%%%%%%%%%%%%%%%%%%%%%%%%%%%%%%%%%%%%
\setlength{\tabcolsep}{0.9em} % for the horizontal padding
%{\renewcommand{\arraystretch}{1.3}% for the vertical padding
%\shortstack \cline{2-7} 

\setlength{\tabcolsep}{0.2em} 
\begin{table*}[ht] \small 
\centering 	
\begin{tabular}{|c|c c c c c c c c c|} 		
\hline    		
\multirow{2}{*}{Depth}	   & \multirow{2}{*}{Method}  &\multirow{2}{*}{\shortstack {Fine-\\tune?}}  &\multirow{2}{*}{\shortstack {Top-1 Accu.\\Baseline(\%)} }  &\multirow{2}{*}{\shortstack {Top-1 Accu.\\Accelerated(\%)} } &\multirow{2}{*}{\shortstack {Top-5 Accu.\\Baseline(\%)} }   &\multirow{2}{*}{\shortstack {Top-5 Accu.\\Accelerated(\%)} }   &\multirow{2}{*}{\shortstack {Top-1 Accu.\\ Drop(\%)} }  &\multirow{2}{*}{\shortstack {Top-5 Accu.\\ Drop(\%)} } & \multirow{2}{*}{\shortstack {Pruned\\FLOPs(\%)}}    \\
         & & & & & & & & & \\ \hline                
\multirow{2}{*}{18} &~\cite{Dong_2017_CVPR} &N   &69.98 &66.33 & 89.24     & 86.94	&3.65	 &2.30   & 34.6		 \\	       		  
%& Ours(30\%) &N &\textbf{70.23$\pm$0.06}	&\textbf{67.25$\pm$0.13} & \textbf{89.51$\pm$0.10}  & \textbf{87.76$\pm$0.06}   &\textbf{2.98}   & \textbf{1.75} & \textbf{41.8} \\\hline       
& Ours(30\%) &N &\textbf{70.28}	&\textbf{67.10} & \textbf{89.63}  & \textbf{87.78}   &\textbf{3.18}   & \textbf{1.85} & \textbf{41.8} \\\hline 

\multirow{3}{*}{34}	          
&~\cite{Dong_2017_CVPR} &N   & 73.42          &\textbf{72.99} 	&91.36	 &\textbf{91.19}	&\textbf{0.43} 	&\textbf{0.17} 	& 24.8	\\         
&~\cite{li2016pruning} &Y    & {73.23}          & 72.17 	&-	 &-	&{1.06} 	&- 	& 24.2	 \\  			  
& Ours(30\%)	&N 	&\textbf{73.92}		&71.83	&\textbf{91.62}   & 90.33   & 2.09  & 1.29 & \textbf{41.1}    \\    \hline     
    
\multirow{4}{*}{50}	&~\cite{He_2017_ICCV} &Y 	&- 	&-	& 92.20    &90.80  	&- 	& 1.40	 &\textbf{50.0}	 	\\     		  
&\cite{Luo_2017_ICCV}  &Y  &72.88 	&72.04  & 91.14   & 90.67              & \textbf{0.84} 	&\textbf{0.47}	& 36.7  \\  		  
& Ours(30\%) & N  &\textbf{76.15}		&\textbf{74.61}		&\textbf{92.87}	  & \textbf{92.06} &1.54	 & 0.81 & 41.8  	 \\
& Ours(30\%) & Y  &\textbf{76.15}		&{62.14}		&\textbf{92.87}	  & {84.60} &14.01	 & 8.27 & 41.8  	 \\\hline  

\multirow{2}{*}{101}	    	& Ours(30\%) & N  &\textbf{77.37}		& 77.03 	&\textbf{93.56}	  &  93.46  & 0.34 	 &  0.10  & \textbf{42.2}    \\  
& Ours(30\%) & Y  &\textbf{77.37}		&\textbf{77.51}		&\textbf{93.56}	  & \textbf{93.71} &\textbf{-0.14}	 & \textbf{-0.20}& \textbf{42.2}    \\ 		
\hline    	
\end{tabular} 	
\caption{
Comparison of pruning ResNet on ImageNet.
``Fine-tune?'' and "Accu. Drop" have the same meaning with Tab.~\ref{table:cifar10_accuracy}.
%``Fine-tune?'' column, ``Y'' and ``N'' indicate whether to use the pre-trained model as initialization or not.
%The "Accu. Drop" is the accuracy of the accelerated model minus that of the baseline model, smaller is better.
%We run ResNet-18 for three times to get the mean and std.
%For the other depths of ResNet, we just list the one-view accuracy.
} 
\label{table:imagenet_accuracy}
\end{table*}

%%%%%%%%%%%%%%%%%%%%%%%%%%%%%%%%%%%%%%%%%%%%%%%%%%%%%%%%%%%%%%%%%%%%%%%%%%%%%%%%%%%%%%%%%%%%%%%%

\begin{table}[ht]
\small
\setlength{\tabcolsep}{0.3em}
\begin{center}
\begin{tabular}{| c | c | c | c  | c |}
\hline
\multirow{2}{*}{Model} & \multirow{2}{*}{\shortstack{Baseline\\time (ms)}} & \multirow{2}{*}{\shortstack {Pruned\\time (ms) }} & \multirow{2}{*}{\shortstack {Realistic \\Speed-up(\%)}} &\multirow{2}{*}{\shortstack {Theoretical\\Speed-up(\%)} }\\
               &         &          &        &          \\ \hline
ResNet-18      & 37.10   &  26.97   & 27.4   &  41.8    \\  
ResNet-34      & 63.97   &  45.14   & 29.4   &  41.1    \\  
ResNet-50      & 135.01  &  94.66   & 29.8   &  41.8    \\ 
ResNet-101     & 219.71  & 148.64   & 32.3   &  42.2    \\ \hline
\end{tabular}
\end{center}
\caption{
Comparison on the theoretical and realistic speedup.
We only count the time consumption of the forward procedure.
%We use the GTX1080 GPU with a batch size of 64.
}
\label{table:Comparison_Speed}
\end{table}
%%%%%%%%%%%%%%%%%%%%%%%%%%%%%%%%%%%%%%%%%%%%%%%%%%%%%%%%%%%%%%%%%%%%%%%%%%%%%%%%%%%%%%%%%%%%%%%%%%%

 \textbf{Settings}.
For ILSVRC-2012 dataset, we test our SFP on ResNet-18, 34, 50 and 101; and we use the same pruning rate 30\% for all the models.
All the convolutional layer of ResNet are pruned with the same pruning rate at the same time.
(We do not prune the projection shortcuts for simplification, which only need negligible time and do not affect the overall cost.)

\textbf{Results.}
Tab.~\ref{table:imagenet_accuracy} shows that SFP outperforms other state-of-the-art methods.
For ResNet-34, SFP without fine-tuning achieves
more inference speedup to the hard pruning method~\cite{Luo_2017_ICCV}, but the accuracy of our pruned model exceeds their model by 2.57\%.
Moreover, for pruning a pre-trained ResNet-101, SFP reduces more than 40\% FLOPs
of the model with even 0.2\% top-5 accuracy increase, which is the
state-of-the-art result. In contrast, the performance degradation is inevitable for hard filter pruning method.
Maintained model capacity of SFP is the main reason for the superior performance. In addition, the non-greedy all-layer pruning method may have a better performance than the locally optimal solution obtained from previous greedy pruning method, which
seems to be another reason. Occasionally, large performance degradation happens for the pre-trained model (e.g., 14.01\% top-1 accuracy drop for ResNet-50). This will be explored in our future work.

To test the realistic speedup ratio, we measure the forward
time of the pruned models on one GTX1080 GPU with a batch size of 64 (shown in Tab.~\ref{table:Comparison_Speed}).
The gap between theoretical and realistic model may come from and the limitation of IO delay, buffer switch and efficiency of BLAS libraries.

%%%%%%%%%%%%%%%%%%%%%%%old version
% For ILSVRC-2012 dataset, we test out SFP on ResNet-18, 34, 50 and
% 101, and for every structure of different depths, we use the same pruning
% rate 30\%. All the convolutional layer of ResNet are pruned with the
% same pruning rate at the same time. \DXYB{We leave the projection shortcut unpruned.} The result are shown in Table~\ref{table:cifar10_accuracy}.

% Our SFP could achieve a better performance than other state-of-the-art
% methods. For example, for ResNet-34, our SFP without fine-tuning achieves
% a similar speed up to \cite{Luo_2017_ICCV}, but the accuracy of
% our pruned model exceeds their model by 3.6\%. Moreover, for pruning a
% pre-trained ResNet-101, our SFP can reduce more than 50\% of FLOPs of the model 
% with even 0.2\% top-5 accuracy increase, which is the state-of-the-art
% result. \DXYB{Maintained
% model capacity during SFP is the main reason for the superior performance.
% In addition, the non-greedy all-layer pruning method may have better performance than the locally
% optimal solution obtained from previous greedy pruning method, which
% seems to be another reason. }

% In order to test the realistic speedup ratio, we measure the forward time of the pruned models on one GTX1080 GPU with batch size 1. The results are shown in Table~\ref{table:Comparison_Speed}. The gap between theoretical and realistic model may come from non-tensor layers and the limitation of IO delay, buffer switch and efficiency of BLAS libraries.

%%%%%%%%%%%%%%%%%%%%%%%%%%%%%%%%%%%%%%%%%%%%%%%%%%%%%%%%%%%%%%%%%%%%%%%%%%%%%%%%%%%%%%%%%%%%%%%%%%%

\subsection{Ablation Study}\label{1-norm and 2-norm}
We conducted extensive ablation studies to further analyze each component of SFP.

\textbf{Filter Selection Criteria.}
The magnitude based criteria such as $\ell_{p}$-norm are widely used to filter selection because computational resources cost is small~\cite{li2016pruning}.
We compare the $\ell_{2}$-norm and $\ell_{1}$-norm. For $\ell_{1}$-norm criteria, the accuracy of the model under pruning rate 10\%, 20\%, 30\% are 93.68$\pm$0.60\%, 93.68$\pm$0.76\% and 93.34$\pm$0.12\%, respectively. While for $\ell_{2}$-norm criteria, the accuracy are 93.89$\pm$0.19\%, 93.93$\pm$0.41\% and 93.38$\pm$0.30\%, respectively.
The performance of $\ell_{2}$-norm criteria is slightly better than that of $\ell_{1}$-norm criteria.
The result of $\ell_{2}$-norm is dominated by the largest element,
while the result of $\ell_{1}$-norm is also largely affected by other small elements.
Therefore, filters with some large weights would be preserved by the $\ell_{2}$-norm criteria. So the corresponding discriminative features are kept so the performance of the pruned model is better.

% \textbf{Filter Selection Criteria.}
% %\label{1-norm and 2-norm}
% Filter selection criteria such as magnitude~\cite{li2016pruning},
% entropy~\cite{luo2017entropy} and Taylor expansion~\cite{molchanov2016pruning}
% have been reported. However, the calculation of entropy and Taylor
% expansion method will cost much more computational resources than
% the magnitude based criteria, so we choose magnitude based criteria.

% We compare the $\ell_{2}$-norm and $\ell_{1}$-norm selection criteria
% and the result is shown in Table~\ref{table:2norm}. We find that
% the performance of $\ell_{2}$-norm criteria is slightly better than that of
% $\ell_{1}$-norm criteria. The result of $\ell_{2}$-norm is dominated by the
% largest element, while the result of $\ell_{1}$-norm is also largely
% affected by other small elements. Therefore, a filter with some large
% weights would be preserved in the $\ell_{2}$-norm criteria. Consequently,
% the corresponding discriminative features are kept so the performance
% of the pruned model is better. 

\iffalse
\begin{table}[!t]
\setlength{\tabcolsep}{0.48em}  
\small 
\centering 	
\begin{tabular}{|c|c|c|c|} 		  \hline
Pruning rate(\%)   &   10             & 20   &30   \\   \hline        
$\ell_1$-norm      & 93.68 $\pm$ 0.60 &93.68 $\pm$ 0.76  &93.34 $\pm$ 0.12 \\ 		
$\ell_2$-norm      & \textbf{93.89 $\pm$ 0.19} &\textbf{93.93 $\pm$ 0.41} &\textbf{93.38 $\pm$ 0.30}\\ 		     \hline
\end{tabular}  	
\caption{
Accuracy of CIFAR-10 on ResNet-110 under different pruning rate with different filter selection criteria.
}
\label{table:2norm}
\end{table}
\fi

\begin{figure}[t]
\center
\subfigure[Different Pruning Rates]{
\label{fig:different_rate}
\includegraphics[width=0.47\linewidth]{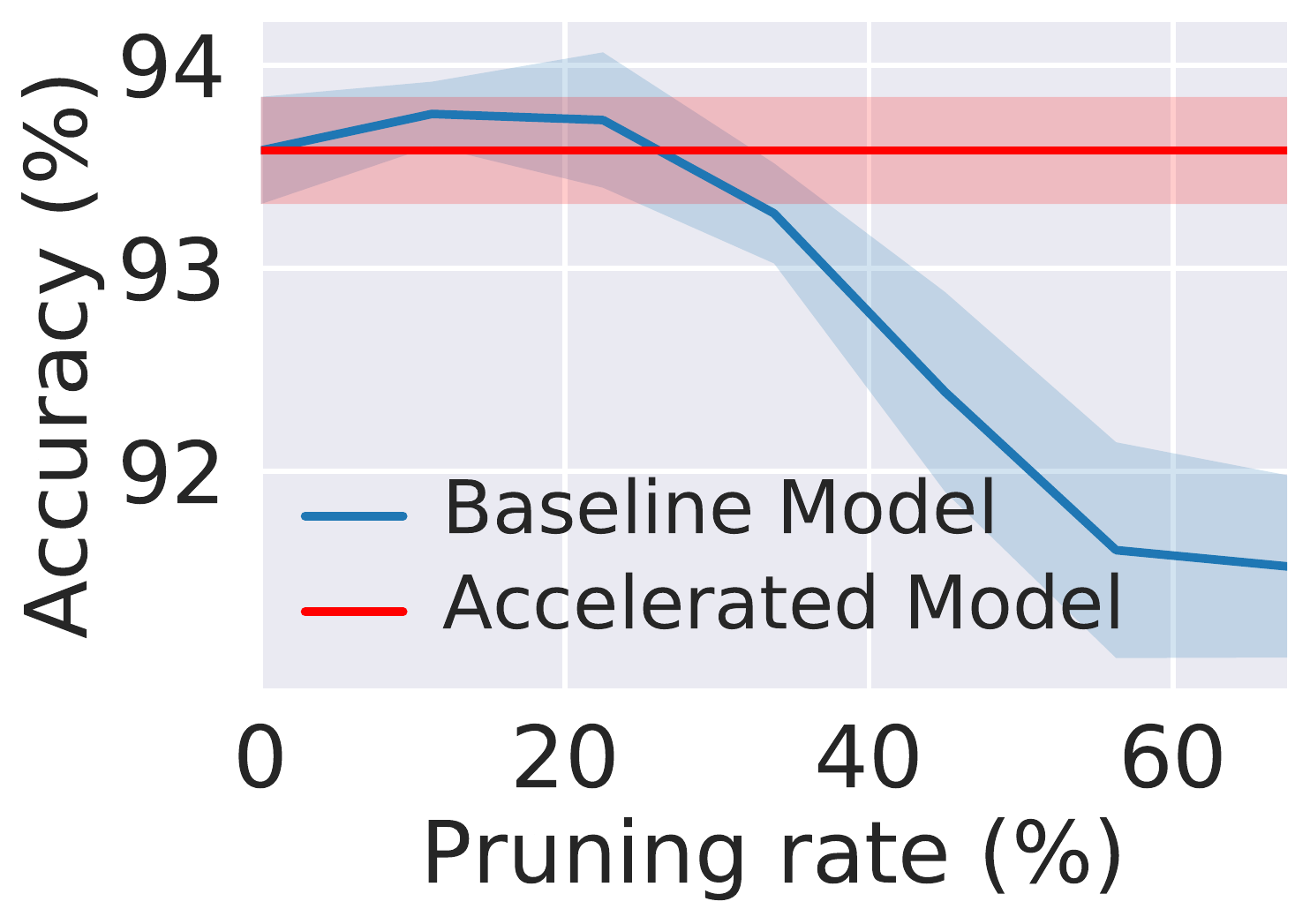}
}
\subfigure[Different SFP Intervals]{
\label{fig:different_epoch}
\includegraphics[width=0.47\linewidth]{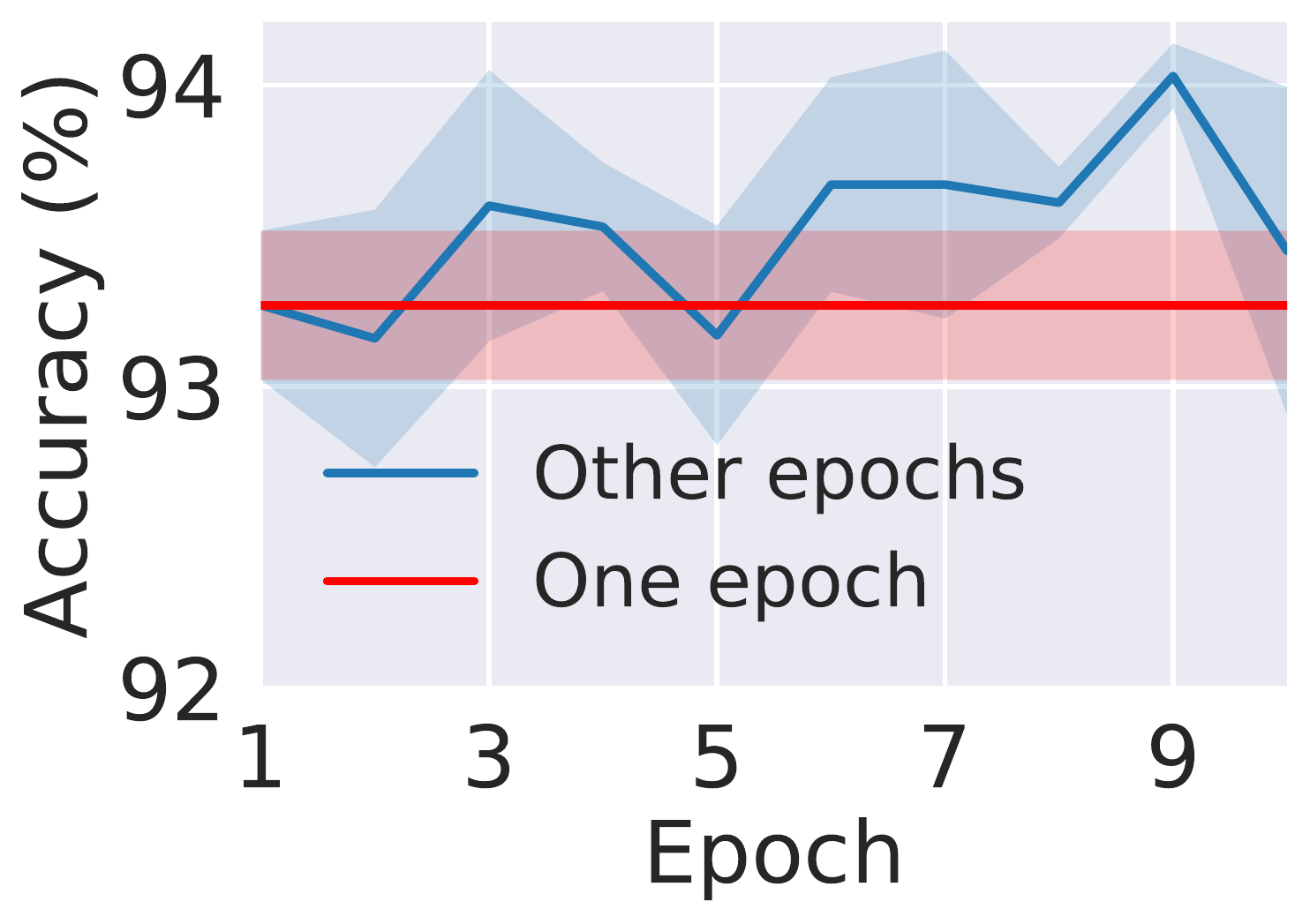}
}
\caption{
Accuracy of ResNet-110 on CIFAR-10 regarding different hyper-parameters. (Solid line and shadow denotes the mean and standard deviation of three experiment, respectively.)
}
\vspace{-2mm}
\label{fig:resnet_cifar100}
\end{figure}

\textbf{Varying pruning rates.}
To comprehensively understand SFP, we test the accuracy of different pruning
rates for ResNet-110, shown in Fig.~\ref{fig:different_rate}.
As the pruning rate increases, the accuracy of the pruned model first rises above the baseline model and then drops approximately linearly.
For the pruning rate between 0\% and about 23\%, the accuracy
of the accelerated model is higher than the baseline model.
This shows that our SFP has a regularization effect on the neural network because
SFP reduces the over-fitting of the model.

\textbf{Sensitivity of SFP interval.}
By default, we conduct our SFP operation at the end of every training
epoch.
However, different SFP intervals may lead to different performance;
so we explore the sensitivity of SFP interval. We use the ResNet-110
under pruning rate 30\% as a baseline, and change the SFP interval
from one epoch to ten epochs, as shown in Fig.~\ref{fig:different_epoch}.
It is shown that the model accuracy has no large fluctuation along
with the different SFP intervals.
Moreover, the model accuracy of most (80\%) intervals surpasses the accuracy of one epoch interval.
Therefore, we can even achieve a better performance if we fine-tune this parameter.

\textbf{Selection of pruned layers.}
Previous works always
prune a portion of the layers of the network. Besides, different layers
always have different pruning rates. For example,~\cite{li2016pruning}
only prunes insensitive layers, ~\cite{Luo_2017_ICCV} skips the
last layer of every block of the ResNet, and~\cite{Luo_2017_ICCV}
prunes more aggressive for shallower layers and prune less for deep
layers. Similarly, we compare the performance of pruning first and
second layer of all basic blocks of ResNet-110. We set the pruning
rate as 30\%. The model with all the first layers of blocks pruned
has an accuracy of $93.96\pm0.13\%$, while that with the
second layers of blocks pruned has an accuracy of $93.38\pm0.44\%$.
Therefore, different layers have different sensitivity for SFP, and careful selection of pruned layers would potentially lead to performance improvement, although more hyper-parameters are needed.

\section{Conclusion and Future Work}

In this paper, we propose a soft filter pruning (SFP) approach to accelerate the deep CNNs.
During the training procedure, SFP allows the pruned filters to be updated.
This soft manner can maintain the model capacity and thus achieve the superior performance.
Remarkably, SFP can achieve the competitive performance compared to the state-of-the-art without the pre-trained model.
Moreover, by leveraging the pre-trained model, SFP achieves a better result and advances the state-of-the-art.
%in the field of model acceleration.
Furthermore, SFP can be combined with other acceleration algorithms, e.g., matrix decomposition and low-precision weights, to further improve the performance.

%%%%%%%%%%%%%%%%%%%%%%%%%%%%%%%%%%%%%%%%%%%%%%%%%%%%%%%%%%%%%%%%%%%%%%%%%%%%%%%%%%%%%%%%%%%%%%%%%%%
{
%\small
\section*{Acknowledgments}
Yi Yang is the recipient of a Google Faculty Research Award.
We acknowledge the Data to Decisions CRC (D2D CRC), the Cooperative Research Centres Programme and ARC's DECRA (project DE170101415) for funding this research. We thank Amazon for the AWS Cloud Credits.
}

%% The file named.bst is a bibliography style file for BibTeX 0.99c
{
\bibliographystyle{named}
\bibliography{ijcai18}
}

\end{document}